 \documentclass[pmlr,twocolumn,10pt]{jmlr} 

\usepackage{booktabs}
\usepackage[load-configurations=version-1]{siunitx} 

\newcommand{\equal}[1]{{\hypersetup{linkcolor=black}\thanks{#1}}}

\theorembodyfont{\upshape}
\theoremheaderfont{\scshape}
\theorempostheader{:}
\theoremsep{\newline}

\jmlrvolume{LEAVE UNSET}
\jmlryear{2021}
\jmlrsubmitted{LEAVE UNSET}
\jmlrpublished{LEAVE UNSET}
\jmlrworkshop{Machine Learning for Health (ML4H) 2021} 

\title[A Machine Learning Approach for Recruitment Prediction in Clinical Trial Design]{A Machine Learning Approach for Recruitment Prediction in Clinical Trial Design}

\author{%
\Name{Jingshu Liu$^1$}\equal{These authors contributed equally and are also corresponding authors} \Email{jingshu.liu@3ds.com}\\
\Name{Patricia J Allen$^2$}\footnotemark[1] \Email{pj.allen@3ds.com}\\
   \Name{Luke Benz$^3$} \Email{lukesbenz@gmail.com}\\
   \Name{Daniel Blickstein$^1$} \Email{daniel.blickstein@3ds.com}\\
   \Name{Evon Okidi$^2$} \Email{evon.okidi@3ds.com}\\
   \Name{Xiao Shi$^1$} \Email{xiao.shi@3ds.com}\\
\addr $^1$350 Hudson Sreet 5F, New York, NY 10014 \\
$^2$110 High Street 6F, Boston, MA 02110 \\
$^3$170 Brookline Ave. Unit 425, Boston, MA 02215
\\[-4ex]
}

\usepackage{url}

\usepackage[singlelinecheck=false ]{caption}
\usepackage{rotating}
 
\begin{document}
\maketitle

\begin{abstract}
Significant advancements have been made in recent years to optimize patient recruitment for clinical trials, however, improved methods for patient recruitment prediction are needed to support trial site selection and to estimate appropriate enrollment timelines in the trial design stage. In this paper, using data from thousands of historical clinical trials, we explore machine learning methods to predict the number of patients enrolled per month at a clinical trial site over the course of a trial's enrollment duration. We show that these methods can reduce the error that is observed with current industry standards and propose opportunities for further improvement.

\end{abstract}

\section{Introduction}
\label{sec:intro}

A key success factor in clinical trials is the timely recruitment of trial participants. A 2020 study of 87 trials from the Tufts Center for the Study of Drug Development suggests that efforts to optimize the recruitment process in recent years have been successful: enrollment timelines were within their target in 77\% of the studies sampled, increased from 48\% in 2012 \citep{tuftsCSDD_2020}. This study, however, was limited in scope to a selected few large pharmaceutical companies, and other recent studies suggest that more precise and granular prediction of patient recruitment in the trial design stage is needed to further reduce the risk of trial recruitment delays and failures \citep{briel_2016,healy_2018}.

Research in this field is often limited by the volume of clinical trial data to generate meaningful enrollment rate estimates and to validate proposed methods. A recent review by \citet{gkioni_2019} described multiple statistical models that have been proposed over the years, however 11 out of the 13 reviewed papers required investigator-provided enrollment rate estimations to generate predictions in the design stage. The remaining two papers either did not validate their methods or only validated on a small sample of studies \citep{bakhshi2013, Gajewski2008}. Additionally, many of these past approaches assume that enrollment rates are constant over time or even across trial sites, which rarely holds in practice. While it is likely that other approaches exist, methods and evaluation results for recruitment prediction in clinical trials are rarely reported.

In this work, we propose a data driven, machine learning (ML) approach to predict site enrollment counts for every month over the course of a prospective clinical trial while accounting for variations in site initiation times, seasonality, and other site and study characteristics. This approach is based on patient enrollment records from thousands of historical clinical trials and enables predictions to be made using only the characteristics of the prospective trial and a list of trial sites. To the best of our knowledge, this is the first reported application of ML for clinical trial enrollment prediction where performance of multiple methods is evaluated and reported based on a large number of trials.

\vspace{-5mm}

\section{Data}
\label{sec:data}

\vspace{-1mm}
\subsection{Data Sources}

\vspace{-1mm}
Medidata's Electronic Data Capture (EDC) system integrates clinical trial operational data from various sponsors\footnote{The sponsor of a clinical trial is the pharmaceutical company that initiated and oversaw the trial.} or contract research organizations (CRO\footnote{A CRO is a company that is optionally contracted to manage clinical trials for a sponsor.}) and disease areas globally. To predict trial enrollment we leverage the operational data at the trial (i.e., ``study"), trial site (i.e., ``study-site") and patient levels, such as 1) trial eCRF finalization date\footnote{eCRF stands for electronic case report form. Its finalization date in the EDC system is used as a proxy for the trial initialization date.}, disease indication\footnote{A disease indication corresponds to one or more ICD-9 disease codes. These  indications are hierarchically combined into indication groups and then into therapeutic areas.}, phase, sponsor and/or CRO; 2) study-site creation date\footnote{This is an EDC system date that captures when a healthcare facility record is added to a trial database, used as a proxy for the study-site activation date which marks the time from which a site can begin enrolling.}, country; 3) patient enrollment status and enrollment date. All patient data has been removed or anonymized, and trial sponsors and/or CROs have granted Medidata authorization to use this combined operational data for research purposes. 

Medidata Acorn AI has invested substantially in standardizing site metadata across trials. By the time the data was generated for this work, about 20K unique healthcare facilities have been identified across over 500K study-sites. This unique site identifier is used to generate features that track site enrollment performance in previous trials, as explained in detail in \sectionref{sec:feature}.

Additionally, public data sources are used to augment the trial operational data, including age and gender inclusion criteria and number of trial arms from the AACT database \citep{ctgov}, and country population and population density data from The World Bank Group \citep{worldbank}.

\subsection{Cohort Summary}
Our initial cohort contains over 11K trials that were initiated between 2008 and 2020 with at least 1 patient enrolled. Trials with potentially erroneous or incomplete data are removed, such as those with abnormally short enrollment duration (e.g., less than 4 months), missing or misaligned key enrollment milestone dates, and/or unknown therapeutic area (TA). Trials with likely bulk-uploaded enrollment data are detected and excluded, such as those with over 90\% of subjects enrolled in the same calendar month. Certain types of studies are excluded because of their distinct enrollment patterns or limited data availability, including observational, pediatric and device trials. \tableref{tab:waterfall} in the appendix summarizes the number of studies, study-sites and patients after each filtering step. 

The final modeling dataset contains about 1.2 million enrolled patients from 7,542 trials and over 270K study-sites. The trials span over 400 sponsors, 80 CROs, 15 TAs, 600 indications and phase I through IV. Oncology is the largest TA (43\% of studies), and phase II and III are the dominating phases (30\% and 43\% of studies, respectively). Over 18K unique healthcare facilities from over 90 countries are represented, with the U.S. being the most represented country (40\% of study-sites). \tableref{tab:datasummary} in the appendix summarises the enrollment related statistics.

\section{Methods}
\label{sec:method}

\subsection{Problem Definition}
\label{sec:def}

\begin{figure}[htbp]
\floatconts
  {fig:nodes}
  {\vspace{-1mm}}
  {\includegraphics[width=1.0\linewidth]{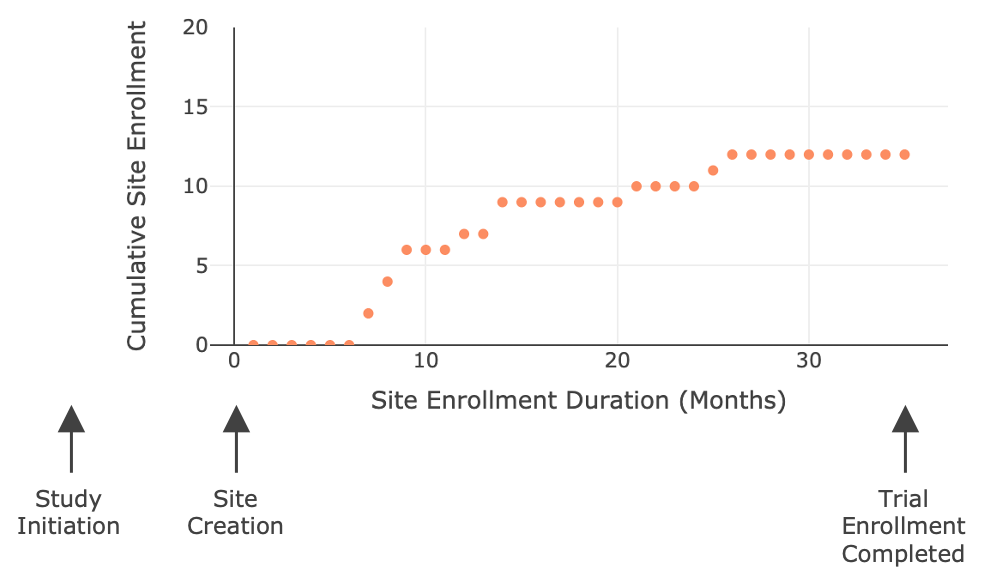}}
  {\vspace{-8mm}}
  {\caption{Example Cumulative Site Enrollment}}
\end{figure}

The aim of the present work is to predict the monthly enrollment count for a clinical trial site throughout the study-site enrollment duration, i.e., from the study-site creation date until the target enrollment for the trial has been reached. These monthly site enrollment count predictions can then be aggregated to estimate country and study level enrollment timelines. Of note, we use the observed study-site creation date for model training and evaluation in the present work for simplicity; additional information regarding the prediction of time to study-site creation is provided in \sectionref{sec:site enrollment start date}.

\subsection{Feature Generation}
\label{sec:feature}
We generate features at the study, country, and site levels, as well as time-varying features at the site-month level, using similar methods to capture performance from relevant historical trials. Relevant historical trials include those that completed enrollment at the same clinical trial site, within the same country, and/or had similar trial characteristics such as the TA or phase. Performance metrics of interest include site enrollment rates, monthly site enrollment counts, site query rates, and other data quality metrics. For example, to compute the mean historical enrollment rate by TA and country for a U.S. site in an oncology trial with an eCRF finalization date of January 10, 2018, we would compute the mean site enrollment rate of all study-sites in the U.S. that completed enrollment in oncology trials prior to January 10, 2018.

Additional features aiming to estimate disease prevalence are also derived using historical trial data. More specifically, the indication disease prevalence is estimated globally and within each country by computing the percentage of patients enrolled in historical trials for a given indication compared to patients enrolled historically across all indications. 

These historical features are often missing due to lack of historical trials within a particular set of grouping variables. In general, we apply various forms of hierarchical imputation that use values from features with higher level groupings to impute missing values for features with lower level groupings. For example, for features that are grouped by indication, we impute missing values using the corresponding value at the indication group level, followed by the TA level if missing values still exist. 

To capture variation over time, we include a number of time related features at the study-site-month level such as the number of months since the study-site creation date, the calendar month and the number of days in a given month. In total, our starting feature set consists of roughly 100 study level, 30 country level, 100 site level and 15 site-month level features.

\subsection{Models}
\noindent\textbf{Baseline Models} We develop two baseline models with constant study-site level enrollment rate prediction for performance comparison. The first uses the mean historical enrollment rate\footnote{Defined as the study-site enrollment count divided by study-site enrollment duration.} observed for a given site within a given indication. If a site lacks history within a given indication, we impute the value hierarchically starting from country and indication to indication, indication group and finally TA. 

The second baseline is an XGBoost \citep{xgb} model trained on site enrollment rate with a tweedie \citep{tweedie} loss function and the 
study, country and site level features as described in Section \ref{sec:feature}. 

\vspace{3mm}

\noindent \textbf{Monthly Enrollment Prediction Models} We explored three methods for this prediction problem: LightGBM \citep{lightGBM} with a tweedie loss function, Zero-Inflated Poisson (ZIP) regression \citep{zip}, and a family of hurdle models \citep{Hurdle} with Poisson, truncated Poisson or negative binomial count distributions. Selection of these methods was motivated primarily by the unique shape of the monthly site enrollment count distribution which has a large mass at zero due to non-enrolling sites. LightGBM offers additional advantages in handling non-linear relationships, interactions and missing values, as well as having a faster training time than other ensembled tree models like XGBoost.

\subsection{Feature Selection and Processing for Monthly Enrollment Prediction Models}
Prior to model training, we apply a set of pre-processing steps, including winsorizing extreme values of the historical features, removing highly correlated features, and pooling categorical features such as indication. For the ZIP and hurdle models, additional steps are required such as imputing missing data, one-hot encoding categorical variables, removing highly sparse variables, and normalizing feature values. The final models contained 90 features,\footnote{About 50 study level, 10 country level, 25 site level features are included in the final models.} and all site-month level features are preserved.  

\section{Result Summary}
The modeling dataset is split randomly by study into discovery and hold-out test sets using a 3:1 ratio. The discovery set is then randomly split by study into 5 cross-validation folds. We conduct model hyper-parameter tuning, including the selection of ZIP and hurdle model distributions with cross-validation. Appendix \ref{app:timesplit} provides discussion on random versus time-based split.

To compare model performance, we evaluate the mean squared error (MSE) and mean absolute error (MAE) of predicted versus observed enrollment at the study, study-site and study-site-month levels. The study level prediction performance is the most important as it drives study enrollment planning and is less affected by noise from the more granular levels. 

\tableref{tab:result} in the appendix shows the detailed cross-validation results across models. The ZIP model has the best study level MAE and MSE, outperforming the historical rate baseline by 53\% and 80\%, respectively. The LightGBM model has the same study level MAE yet worse MSE compared to ZIP. The XGBoost site enrollment rate baseline has the best study-site-month level MAE. However, it tends to under-predict the enrollment rate, leading to inferior study level performance. 

A secondary consideration is the model training and prediction time. The ZIP model takes over a day to train while the LightGBM model only takes a few minutes. Given  comparable model performance, the LightGBM model is selected as the final model for evaluation on the held-out test set.

 \tableref{tab:result1} shows the study level MAE by cross-validation and on the hold-out set. We also generate the MAE of study enrollment on two interim enrollment milestone dates: 50\% of patients enrolled, and 90\% of patients enrolled, on the test set. A large improvement is observed over baselines at each milestone. Additionally, the MSE and MAE at all three levels in the hold-out set are consistent with the cross-validation results for LightGBM.  \\

\begin{table}
\floatconts
  {tab:result1}%
  {\caption{\footnotesize{Model Performance Comparison (Simplified). Cross-validation and hold-out test set results of selected models. Test set results include the MAE of total study enrollment at two interim enrollment milestone dates: 50\% of patients enrolled (PE), and 90\% of patients enrolled, in addition to MAE measured at the last patient enrolled date. Given running time concerns, ZIP and hurdle models performance is not evaluated on the test set.}
  }}%
  
  \resizebox{\linewidth}{!}{
  {\begin{tabular}{|l|c|c|c|c|c|}
        \hline
  Study Level MAE & CV Results & \multicolumn{3}{c|}{Hold-out Test Set Results}   \\ \hline
  \textbf{Models}  & {\textbf{Last PE}} & \textbf{50\% PE} & \textbf{90\% PE}  & {\textbf{Last PE}} \\ \hline
   
   Mean historical enrollment rate  & 136 & 63 & 104 & 120 \\ \hline
   XGB site enrollment rate  & 93  & 51 & 93 & 98 \\ \hline
   \hline
   
   LightGBM tweedie  &\textbf{67} & \textbf{38} & \textbf{61} & \textbf{64}  \\ \hline
   Hurdle (Poisson) & 84  & \_ & \_ & \_ \\ \hline
   Zero-Inflated Poisson  & \textbf{67} & \_ & \_ & \_ \\ \hline
   \hline
  \end{tabular}}}
\end{table}

\noindent \textbf{High Impact Features}
We evaluate feature importance in the LightGBM model using the average gain associated with each feature. The top 5 features are trial indication group, average historical enrollment by month within the same country or TA, site country and the number of months since enrollment started. The relative importance of the top 5 features accounts for 42\% among all features. Appendix \ref{app:f_importance} shows the relative feature importance of the top 20 features. 

\vspace{-1mm}
\section{Discussion and Future Work}
In this work, we show the power of machine learning models for clinical trial enrollment prediction. Given a rich set of trial operational data, we are able to evaluate a range of machine learning models that generate time-varying predictions at the study-site-month level for prospective clinical trials, largely outperforming the more commonly used projections of site enrollment rate. These predictions can serve as guidelines in the trial design stage and can be refined with subject matter expert input and observed enrollment as the trial progresses.


Despite the volume and granularity of the data used, the present work does not account for all factors that impact the complex process of trial recruitment. We discuss some prominent limitations of the current work and areas for development in Appendix \ref{app:future_work}.

\acks{We thank our colleagues from the Data Operations team and Data Science Engineering team at Medidata Acorn AI for their efforts in data standardization, and building and maintaining the data pipeline behind the modeling dataset. We would also like to thank Jef Benbanaste and Anthony Ford for their review and comments that greatly improved the manuscript. Additionally, we thank Hrishikesh Karvir, Fanyi Zhang and Laura Katz for their contributions to an earlier version of this work.

Medidata's analyses are based upon anonymized and aggregated clinical trial data developed pursuant to customer authorization. Medidata's cross-industry cohort is used only under these protections; collaborations with Medidata would apply similar aggregation requirements.

}

\bibliography{ref}
\appendix

\section{Future Work}
\label{app:future_work}
Prominent limitations of the current work and areas for development continued from the main text. 
\subsection{Site Activation Date}
\label{sec:site enrollment start date}
As mentioned in Section \ref{sec:def}, we assume the study-site creation date is known when training the model and reporting model performance in this work. For trial planning, a separate model has been developed to predict the time from the trial initialization date to each site's creation date which can be used in combination with the monthly enrollment predictions. Study level enrollment prediction performance with predicted site creation date is interestingly almost identical to that with the observed site creation date. 

Separately, the actual site activation date is heavily impacted by country regulations and a trial sponsor's planned activation schedule which is not explicitly captured in our historical performance data. We are exploring other systems that better capture site activation events to gain more understanding of this process and further improve our predictions in the trial design stage.

\subsection{Additional Features to Explore}
While the data explored in this work provide useful information about enrollment patterns in historical trials, additional information about patient availability near clinical trial sites, including information about competing trials that are recruiting similar patients nearby, are of interest for further model development. Similarly, data about epidemiological events such as COVID-19 and associated local regulations near sites could further improve model performance under such conditions. In addition, better usage of trial design information, e.g., disease severity or line of therapy, could help to account for enrollment differences across trials.  

\subsection{Prediction Interval Generation}
In addition to the predicted mean enrollment over time, prediction intervals are needed to evaluate uncertainty in the enrollment forecast. Common methods used in the industry are Monte-Carlo simulation \citep{ABBAS2007220} or analytical solutions based on certain underlying stochastic process assumptions \citep{ani2011}. However, we find that a simulation approach under the tweedie distribution assumption has poor calibration results and requires the dispersion parameter to be scaled up substantially to capture uncertainties in the data. Additionally, the simulation approach is relatively slow, and it can become exponentially complex to account for different levels of uncertainty (e.g., the elapsed time on the trial) and/or the covariance structure of sites within the same study. As a result, we are considering data driven approaches such as quantile regression to alleviate these issues.

\section{Challenges and Performance with Time-based Split}
\label{app:timesplit}
For modeling data with a time component, using a time-based training, validation and test data split often provides the most realistic estimation of model performance when applied to new data. However, the trial enrollment data used in the present work poses special challenges to using a time-based split. First of all, trials can enroll for multiple years, and data split by either the enrollment start time or end time would inevitably bias the test set with either shorter or longer enrolling studies. Secondly, trial enrollment duration is often correlated with the therapeutic area; for example, oncology trials often take years to enroll while dermatology trials typically finish recruitment within a year. Thus, in addition to a disproportional enrollment duration distribution between training, validation and test datasets, there would be bias in the trial disease area. The above concerns could be alleviated given a sufficiently large number of historical trials, however, the majority of studies in our modeling dataset started between 2014 and 2019 which is a relative short period to split 3-ways given an average 2-year enrollment duration. Therefore, we use a random split by study to avoid the potential data distribution bias and its impact on model and feature selection decisions.

To understand model performance on future studies, we conduct the following analysis to compare performance under the time-based split and the random split with the selected LightGBM model, parameters and features. For the time-based split, we train a model on historical trials with a trial initiation date before a given date and evaluate model performance on trials that started in the following quarter. We conduct the above process iteratively for each quarter between 2012 and 2020. As a comparison, we use the cross-validation out-of-fold prediction in the same set of studies, where studies are split randomly into the cross-validation folds. Not surprisingly, we observe a study-level MAE increase of about 10 patients prior to 2015 using a time-based split due to only having a few studies in the training set. The gap gradually reduces to less than 2 patients beyond 2018. While the performance metrics become worse under the time-based split, the performance difference between the LightGBM model and the historical rate baseline remains substantial.

\section{Feature Importance}
\label{app:f_importance}
\begin{figure}[h]
\floatconts
  {fig:f_importance}
  {\vspace{-0.5cm}}
  {\includegraphics[width=2.0\linewidth]{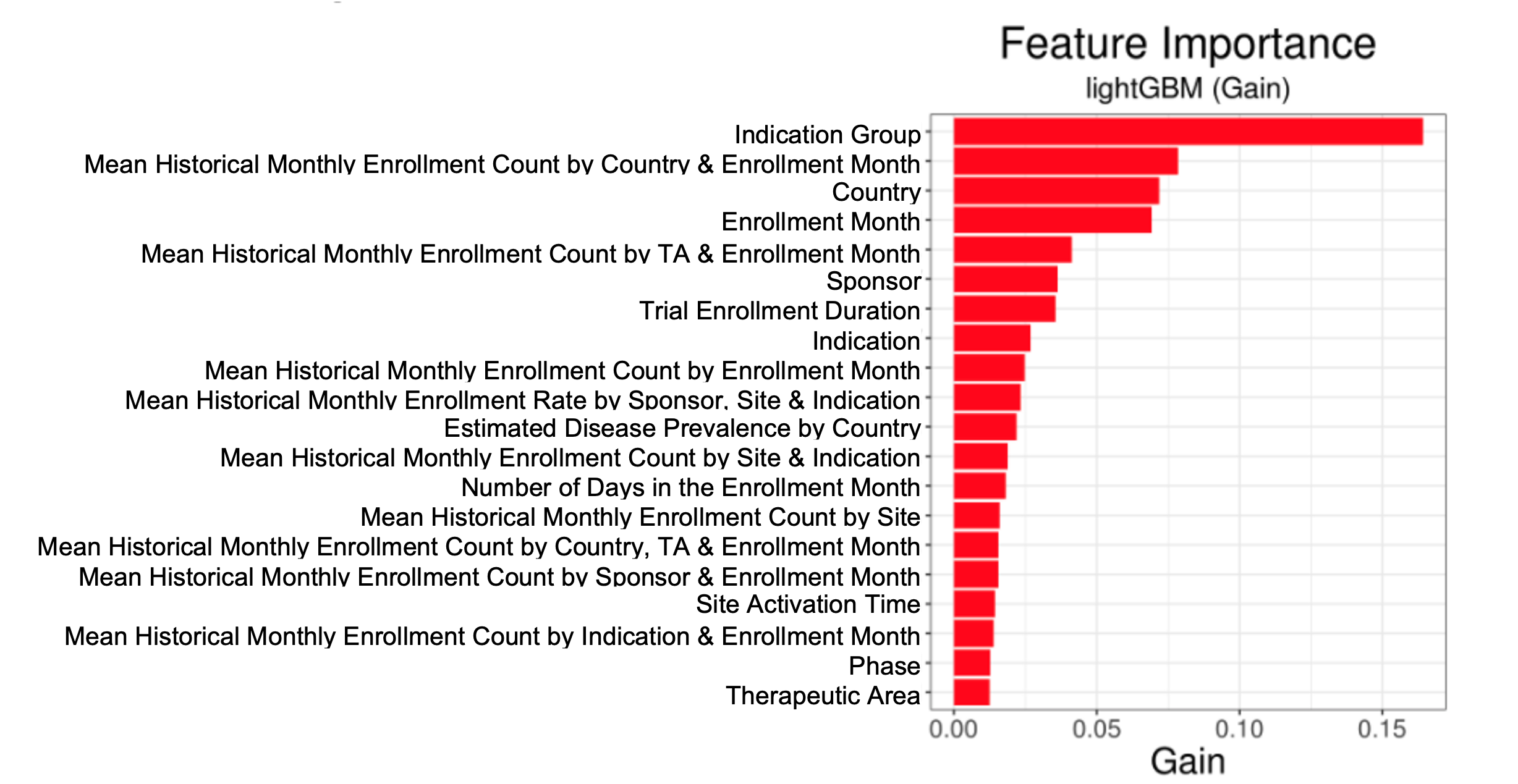}}
\end{figure}

\section{Additional Tables}

\tableref{tab:waterfall} shows the step-wise cohort filtering criteria and the resulting change in the data size. \tableref{tab:datasummary} shows enrollment related summary statistics across trials in the modeling data. \tableref{tab:result} provides more detailed cross-validation results.

\begin{table*}[htbp]
\floatconts
  {tab:waterfall}%
  {\caption{Cohort Summary}}
    {\begin{tabular}{|c|c|c|l|}
        \hline
   \textbf{Studies}  & \textbf{Study Sites} & \textbf{Subjects} & \textbf{Exclusion Criteria} \\\hline
   11,959 & 695,610 & 3,780,709 & None\\ \hline
   9,000 & 310,908 & 2,228,487 & Potentially erroneous or incomplete data \\ \hline
   8,795 & 306,865 & 2,178,280 & Pediatric trials \\ \hline
   8,021 & 295,228 & 1,328,420 & Likely bulk-uploaded enrollment data \\ \hline
   7,566 & 277,301 & 1,210,011 & Observational studies \\ \hline
   7,542 & 276,203 & 1,198,923 & Device studies \\ \hline
    \end{tabular}}
\end{table*}

\begin{table*}
\floatconts
  {tab:datasummary}%
  {\caption{Modeling Data Enrollment Statistics Summary. Study-sites with enrollment rate in the top 2.5\% are viewed as outliers and are removed during data processing.}}%
  {\begin{tabular}{|l|c|c|c|c|c|c|c|}
        \hline
     & \textbf{Mean} & \textbf{SD} & \textbf{Min} & \textbf{25pct} & \textbf{Median} & \textbf{75pct} & \textbf{Max}   \\ \hline
   Trial Enrollment Duration (Months) & 21.0 & 13.5 & 1 & 11 & 18 & 27 & 138\\ 
     Trial Total Enrollment Count & 159.4 & 419.7 & 1 & 15 & 47 & 153  & 12079 \\
    Study-site Enrollment Duration (Months) & 21.4 & 13.1 & $<$1 month & 12 & 19 & 27 & 138\\
    Study-site Total Enrollment Count & 4.3 & 6.7 & 0 & 0 & 2 & 6 & 159 \\
    Study-site-month Enrollment Count & 0.20 & 0.74 & 0 & 0 & 0 & 0 & 52 \\
    Percentage of Non-enrolling Sites within a Trial & 23\% & 24\% & 0\% & 0\% & 18\% & 37\% & 98\%\\ \hline
  \end{tabular}}
\end{table*}

\begin{table*}
\floatconts
  {tab:result}%
  {\caption{Model Performance Comparison - Cross-Validation Results. Both the LightGBM and the ZIP model have the best performance by multiple metrics including study level MAE, study-site level MSE and study-site-month level MSE. The ZIP model additionally has the best study level MSE. The XGBoost site enrollment rate baseline has the best study-site level MAE and study-site-month level MAE. However, the fact that both the study-site level and study-site-month level enrollment distributions have a mass at zero favors models that tend to under-predict. This is the case for the XGBoost site enrollment rate model which leads to inferior study level performance. The standard error of each metric is shown in parentheses}}%
  {\begin{tabular}{|l|c|c|c|c|c|c|}
        \hline
   & \multicolumn{2}{c|}{Study Level} & \multicolumn{2}{c|}{Study-Site Level} &  \multicolumn{2}{c|}{Study-Site-Month Level}  \\ \hline
   \textbf{Models}  & \textbf{MAE} & \textbf{MSE} & \textbf{MAE} & \textbf{MSE} & \textbf{MAE} & \textbf{MSE} \\ \hline
   
   Mean historical enrollment rate & 136 (4.5) & 135357 (12735) & 5.4 (0.013) & 63 (0.4) & 0.44 (3e-4) & 0.55 (1.6e-3) \\ \hline
   XGB site enrollment rate & 93 (4.0) & 101335 (21709) & \textbf{3.5} (0.012) & 41 (0.43) & \textbf{0.27} (3e-4) & 0.51 (1.7e-3) \\ \hline
   \hline
   
   LightGBM tweedie & \textbf{67} (2.7) & 45965 (9475) & 3.6 (0.010) & \textbf{34} (0.37) & 0.32 (3e-4) & \textbf{0.49} (1.6e-3) \\ \hline
   Hurdle (Poisson) & 84 (2.5) & 42395 (6246) & 4.1 (0.010) & 36 (0.34) & 0.37 (3e-4) & 0.50 (1.6e-3) \\ \hline
   Zero-Inflated Poisson & \textbf{67} (1.9) & \textbf{25733} (4789) & 3.7 (0.010) & \textbf{34} (0.35) & 0.34 (3e-4) & \textbf{0.49} (1.6e-3)\\ \hline
   \hline
   
  \end{tabular}}
\end{table*}


\end{document}